\documentclass{article}
\usepackage{caption}
\usepackage{spconf,amsmath,graphicx}
\usepackage{subcaption}  

\usepackage{enumitem}
\setlist{nosep, leftmargin=14pt}

\usepackage{mwe} 
\usepackage{spconf,amsmath,graphicx}
\usepackage{amssymb}
\usepackage{makecell}
\usepackage{booktabs} 
\usepackage{tabularx}
\usepackage[colorlinks=true,
            linkcolor=blue,
            citecolor=blue,
            urlcolor=blue]{hyperref}
            
\setlist{nosep,leftmargin=*}

\title{Self-Learned Representation-Guided Latent Diffusion Model for Breast Cancer Classification in Deep Ultraviolet Whole Surface Images}
\name{
\begin{tabular}{c}
Pouya Afshin$^{1}$, David Helminiak$^{2}$, Tianling Niu$^{3}$, Julie M. Jorns$^{4}$, Tina Yen$^{5}$, Bing Yu$^{3}$, Dong Hye Ye$^{1\dag}$
\end{tabular}
}

\address{
$^{1}$Department of Computer Science, Georgia State University, Atlanta, USA \\
$^{2}$Department of Electrical and Computer Engineering, Marquette University, Milwaukee, USA \\
$^{3}$Joint Dept. of Biomedical Eng., Marquette Univ. and Med. Coll. of Wisconsin, Milwaukee, USA \\
$^{4}$Department of Pathology, Medical College of Wisconsin \\
$^{5}$Department of Surgery, Medical College of Wisconsin \\
}

\begin{document}
%
\maketitle
\begin{abstract}
Breast-Conserving Surgery (BCS) requires precise intraoperative margin assessment to preserve healthy tissue. Deep Ultraviolet Fluorescence Scanning Microscopy (DUV-FSM) offers rapid, high-resolution surface imaging for this purpose; however, the scarcity of annotated DUV data hinders the training of robust deep learning models. To address this, we propose an Self-Supervised Learning (SSL)-guided Latent Diffusion Model (LDM) to generate high-quality synthetic training patches. By guiding the LDM with embeddings from a fine-tuned DINO teacher, we inject rich semantic details of cellular structures into the synthetic data. We combine real and synthetic patches to fine-tune a Vision Transformer (ViT) and use patch-prediction aggregation for WSI-level classification. Experiments using 5-fold cross-validation demonstrate that our method achieves 96.47\% accuracy and reduces the FID score to 45.72, significantly outperforming class-conditioned baselines.

\end{abstract}
\section{Introduction}
\label{sec:intro}
Breast cancer affects one in eight women in the United States, making Breast-Conserving Surgery (BCS) a critical procedure for removing tumors while preserving healthy tissue~\cite{b1,b4}. Deep Ultraviolet Fluorescence Scanning Microscope (DUV-FSM) produces high-resolution, real-time images of tissue surfaces, enabling easier distinction between cancerous and healthy areas. When combined with automated classification, it can help improve margin assessment and reduce extra surgeries. Classifying Deep Ultraviolet Whole Surface Images (DUV WSIs) at full resolution is impractical and downsampling can lose important details. To address this, patch-level Deep Learning (DL) methods have been used ~\cite{b1,b4}. Despite their benefits, deep learning models often don't perform well with limited data, which is common in medical imaging, such as DUV. This shows the need for good data augmentation and stronger learning methods to improve their performance ~\cite{b5, b6}.

While Generative Adversarial Networks (GANs)~\cite{b7} are widely used for generating synthetic data, they often produce limited diversity and suffer from mode collapse. This is particularly problematic for unbalanced DUV WSI patch-level datasets that require fine-grained malignant samples~\cite{b5,b8,b12}. Denoising Diffusion Probabilistic Models (DDPMs)~\cite{b22} offer higher stability and image quality, especially for DUV WSI data~\cite{b5}, but entail high computational costs~\cite{b13}. LDMs~\cite{b15} reduce computation by performing diffusion in a lower-dimensional latent space. Their cross-attention module enables guided image generation using different forms of conditioning information~\cite{b15}. Recent works~\cite{b16,b17} have shown that conditioning diffusion models on Self-Supervised Learning (SSL) embeddings, rather than text or simple class labels, yields semantically richer images. 

\begin{figure*}[ht] 
    \centering
    \includegraphics[width=0.975\textwidth]{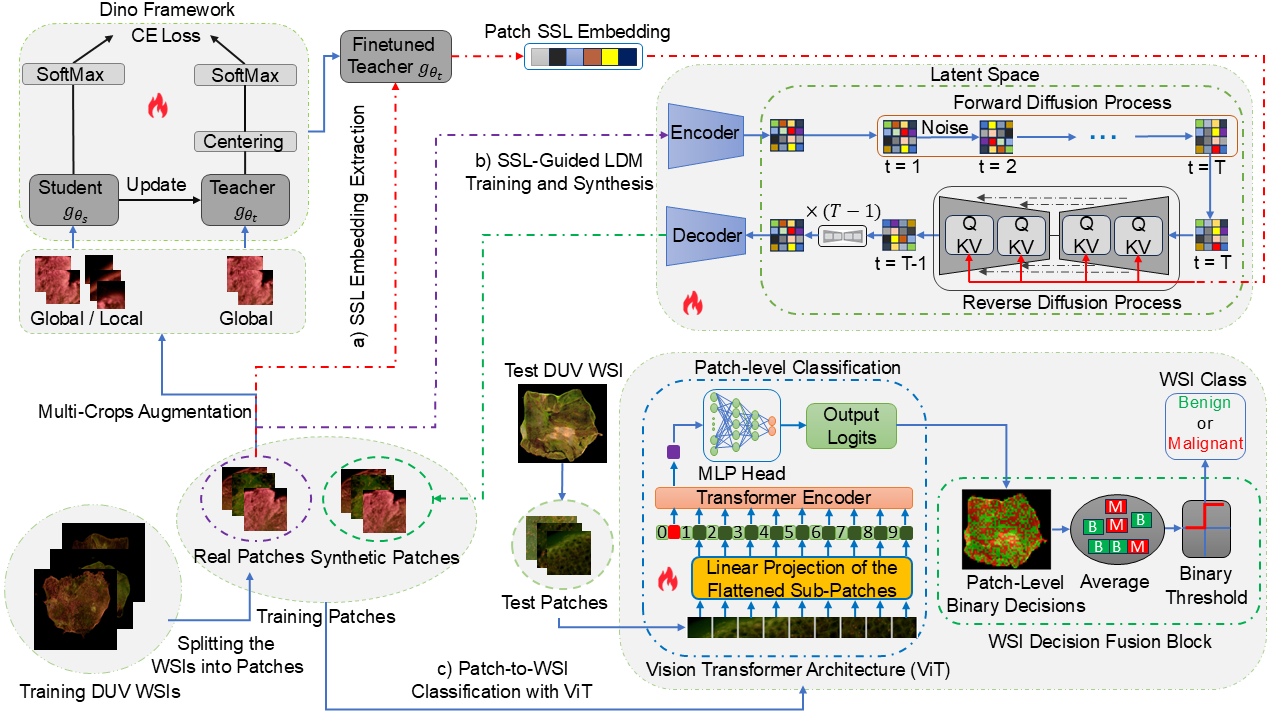} 
\caption{System model overview for DUV WSI classification with SSL-guided LDM augmentation. (a) Real training patches are encoded into SSL embeddings using a fine-tuned DINO teacher model. (b) LDM uses these embeddings to guide its U-Net denoiser via cross-attention during denoising, generating realistic synthetic patches to enrich the training set. (c) A ViT is fine-tuned on both real and synthetic patches for patch-level classification, and WSI-level predictions are obtained by aggregating patch predictions and applying a binary threshold.}
    \label{System-model}
\end{figure*}
SSL-guided LDM has been applied to histopathology and satellite imagery \cite{b17}, but not to DUV WSIs, which exhibit complex features. In this study, inspired by \cite{b17}, SSL embeddings are extracted from a fine‑tuned DINO model \cite{b21} and used to guide an LDM in generating synthetic patches of DUV images. {Detailed tissue structures, cell-level patterns, and cancer morphologies are captured by these embeddings, which offer richer and more discriminative information than standard class labels or other self‑supervised embeddings \cite{b21}}. The synthetic patches are combined with real training patches to fine-tune a pretrained Vision Transformer (ViT)~\cite{b3}, with patch-level predictions aggregated for WSI classification. In summary, this study makes the following contributions:
\begin{itemize}
    \item \textbf{SSL-Guided Patch Generation using LDM}:DINO-based SSL embeddings guide an LDM to generate realistic DUV WSI patches.
    
    \item \textbf{Enhanced DUV WSI Classification:}
    WSI-level classification performance is improved by augmenting the dataset with synthetic patches generated by the SSL-guided LDM.
\end{itemize}
\section{Methodology}
Generating realistic DUV WSI patches and improving WSI-level classification begins with extracting SSL embeddings from DUV WSI patches using a fine-tuned DINO teacher model. An LDM uses these embeddings as guidance to produce high-quality synthetic patches. Both real and synthetic patches are used to fine-tune a pre-trained ViT, with WSI labels determined by combining patch predictions, and applying a binary threshold.
\subsection{SSL Embedding Extraction}
DINO \cite{b21} is a self-supervised, label-free method that learns rich semantic features for tasks, such as classification or segmentation. It uses a student-teacher framework, where the student learns from multiple augmented views (global and local) to match the teacher’s output from broader, global views. The teacher tracks the student model parameters and provides a smooth and stable reference, via an Exponential Moving Average (EMA), for their training by gradient descent. Sharpening and balancing the teacher’s output prevents collapse and produces useful, discriminative features \cite{b21}.

Let $\mathcal{X} = \{x_i\}_{i=1}^{M}$ be the set of DUV WSIs. Each WSI $x_i$ is split into non-overlapping $400 \times 400$ patches, 
denoted by $\mathbf{p}_i^{j} \in \mathbb{R}^{H \times W \times C}$, where $j = 1, \dots, J$ indexes the patches of the $i$-th WSI. Background-dominated patches are removed, leaving only informative regions \cite{b1}. The patches are used to fine-tune the pre-trained DINO model on DUV WSIs. The student network with parameters $\theta_s$ is trained to match the teacher’s output by minimizing the cross-entropy loss \cite{b21}:
\begin{align}
 \min_{\theta_s} \sum_{\mathbf{p}_i^{j} \in \mathcal{T}} 
\sum_{\substack{\mathbf{p'}_i^{j} \in \mathcal{S} \\ \mathbf{p'}_i^{j} \neq \mathbf{p}_i^{j}}} 
H\Big(
    \mathbf{P}_t(\mathbf{p}_i^{j}), 
    \mathbf{P}_s(\mathbf{p'}_i^{j})
\Big)
\end{align}
$\mathcal{T}$ and $\mathcal{S}$ are teacher (global) and student (local and global) patch view sets, where $H(a, b) = -a \log b$. The respective outputs, $\mathbf{P}_t$ and $\mathbf{P}_s$, are probability distributions over $V$ dimensions, computed by a temperature-scaled softmax~\cite{b21}.
\begin{align}
\mathbf{P}_s(\mathbf{p}_i^{j})= 
\frac{\exp(g_{\theta_{s}}(\mathbf{p}_i^{j})/\tau_s)}
{\sum_{v=1}^{V} \exp(g_{\theta_{s}}(\mathbf{p}_i^{j})^v/\tau_s)},
\end{align}
Similarly, $\mathbf{P}_t(\mathbf{p}_i^{j})$ uses temperature $\tau_t$ to control output sharpness. As mentioned in \cite{b21}, the teacher network performs better during training; therefore, it is frozen in this study and used to extract SSL embeddings. 

\subsection{SSL-Guided LDM Training and Synthesis}
To synthesize realistic DUV WSI patches, an LDM is applied that leverages the DINO SSL embeddings as semantic guidance. The LDM has two main parts: a Variational Autoencoder (VAE) that maps DUV WSI patches into low-dimensional latent vectors, which are then corrupted by noise, and a U-Net denoiser trained to recover these latent vectors from Gaussian noise using a conditioning signal ~\cite{b17}. To train the LDM, patches $\mathbf{p}_i^{j}$ are resized to $256 \times 256$ and encoded into latent vectors via an encoder $\varepsilon$~\cite{b15}:
\begin{align}
\mathbf{z}_i^{j,0} = \varepsilon(\mathbf{p}_i^{j}) \in \mathbb{R}^{h \times w \times d},
\end{align}
where $\mathbf{z}_i^{j,0}$ is the noise-free latent at time step $t=0$, with height $h$, width $w$, and depth $d$ computed by scaling the original patch dimensions by a downsampling factor $f = H/h = W/w$. A forward diffusion process gradually adds Gaussian noise over \(T\) steps to latent vectors \cite{b15, b20}:
\begin{align}
q(\mathbf{z}_i^{j,t}|\mathbf{z}_i^{j,0}) = 
\mathcal{N}\big(\mathbf{z}_i^{j,t}; \sqrt{\bar{\alpha}_t}\mathbf{z}_i^{j,0}, (1-\bar{\alpha}_t)\mathbf{I}\big).
\end{align}
Here, $\alpha_t = 1 - \beta_t$ indicates how much latent content is kept at each time step, $\beta_{t}$ controls the noise added, and $\bar{\alpha}_t = \prod_{i=1}^t \alpha_i$. The trainable U-Net denoiser $\epsilon_\theta(\mathbf{z}_i^{j,t}, t, \mathbf{y}_i^j)$ predicts the noise added at each diffusion step, guided by SSL embeddings 
$\mathbf{y}_i^j = g_{\theta_t}(\mathbf{p}_i^{j})$ from the fine-tuned DINO teacher encoder, capturing domain-specific DUV features. 
These embeddings guide the U-Net denoiser using cross-attention layers. The LDM denoising loss is defined as \cite{b15, b20}:
\begin{align}
\mathcal{L}_{\mathrm{LDM}} = 
\mathbb{E}_{\mathbf{z}_i^{j,0}, t, \epsilon \sim \mathcal{N}(0,1)} 
\Big[ \lVert \epsilon - \epsilon_\theta(\mathbf{z}_i^{j,t}, t, \mathbf{y}_i^j) \rVert_2^2 \Big],
\end{align}
where $\mathbf{z}_i^{j,t}$, $t$, $\epsilon$, and $\mathbf{y}_i^j$ are the noisy latent, timestep, Gaussian noise, and conditioning embedding. {During synthesis, noise is gradually removed to recover $\mathbf{\hat{z}}_i^{j,0} \sim p(\mathbf{z}_i^{j,0})$, which is then decoded into a synthetic patch by decoder } \(\mathcal{D}\) \cite{b15}:
\begin{align}
\hat{\mathbf{p}}_i^{j} = \mathcal{D}(\mathbf{\hat{z}}_i^{j,0}), \qquad 
\hat{\mathbf{p}}_i^{j} \in \mathbb{R}^{H \times W \times C},
\end{align}
where $\hat{\mathbf{p}}_i^j$ is the reconstructed $j$-th patch from the WSI $x_i$.
\vspace{-5pt}
\subsection{Patch-to-WSI Classification with ViT}
Given the generated patches from LDM, synthetic patches are added to the training set to fine-tune a pre-trained ViT and improve DUV WSI classification. Patches are indexed by $j = 1, 2, ..., L$, where the first $J$ are real patches of WSI $x_i$ and the remaining are synthetic. Patches are resized to $224 \times 224$, for model compatibility. Each patch is split into $N$ smaller sub-patches $\mathbf{s}_i^{jk}$, reshaped and projected using a trainable matrix \( \mathbf{E} \in \mathbb{R}^{(P^2 \cdot C) \times D} \) to obtain its embeddings $\hat{\mathbf{s}}_i^{jk}  \in \mathbb{R}^{1 \times D}$. A trainable class token \( \mathbf{p}_i^{j \text{class}} \in \mathbb{R}^{1 \times D} \) is added at the beginning as a global patch representation. Positional embeddings \( \mathbf{E}_i^{j \text{pos}} \in \mathbb{R}^{(N+1) \times D} \), containing sub-patch location data, are added to form the input sequence \cite{b3}:
\begin{align}
\mathbf{h}_0 = [\mathbf{p}_i^{j \text{class}}, \hat{\mathbf{s}}_i^{j1}, \dots, \hat{\mathbf{s}}_i^{jN}] + \mathbf{E}_i^{j\text{pos}}.
\end{align}
The input $\mathbf{h}_0$ is processed by the transformer encoder through L layers, each with normalization, self-attention, and MLP blocks. The final class token $\mathbf{p}_i^{j \text{class}}$ is normalized and passed to a classification head to predict the patch label \cite{b3}:
\begin{align}
\hat{y}_i^j = \text{MLP}_{\text{head}}(\text{LN}(\mathbf{p}_i^{j \text{class}}) )\in \{0, 1\}.
\end{align}
WSI-level classification is obtained by aggregating the patch predictions. Let $J$ be the number of real patches in WSI $x_i$. The WSI is classified as malignant if the mean of its patch predictions exceeds a threshold $\theta$ and benign otherwise:
\begin{align}
\hat{y}_i =
\begin{cases}
1, & \text{if } \frac{1}{J} \sum_{j=1}^{J} \hat{y}_i^j > \theta, \\
0, & \text{otherwise}.
\end{cases}
\end{align}

\section{Experiment}
In this study, an LDM, guided by SSL embeddings from the DINO teacher model \cite{b21}, was used to generate synthetic patches to improve breast cancer DUV WSI classification. Synthetic patches were generated for both benign and malignant classes to augment the training data, with 5,000 patches per class. SSL embeddings were randomly sampled from the real training patches. {During LDM training, the embedding was randomly dropped $10\%$ of the time to enable classifier‑free guidance in the synthesis model}. Patch-level classification was done with ViT-B/16 using 5-fold cross-validation. {The split was at the WSI level to avoid data leakage. For each fold, one WSI set served as the test set, and the patches from the other WSIs were allocated 80\% for training and 20\% for validation \cite{b3}. The ViT was trained for 40 epochs using SGD with a learning rate of $3\times10^{-4}$ WSI predictions were obtained by aggregating patch predictions with  $\theta = 0.2$}. For comparison, class-conditioned LDM, DDPM, and a baseline without synthesis were evaluated.
\begin{figure*}[ht]
    \centering
    \begin{subfigure}[b]{0.33\textwidth}
        \centering
        \includegraphics[width=\textwidth]{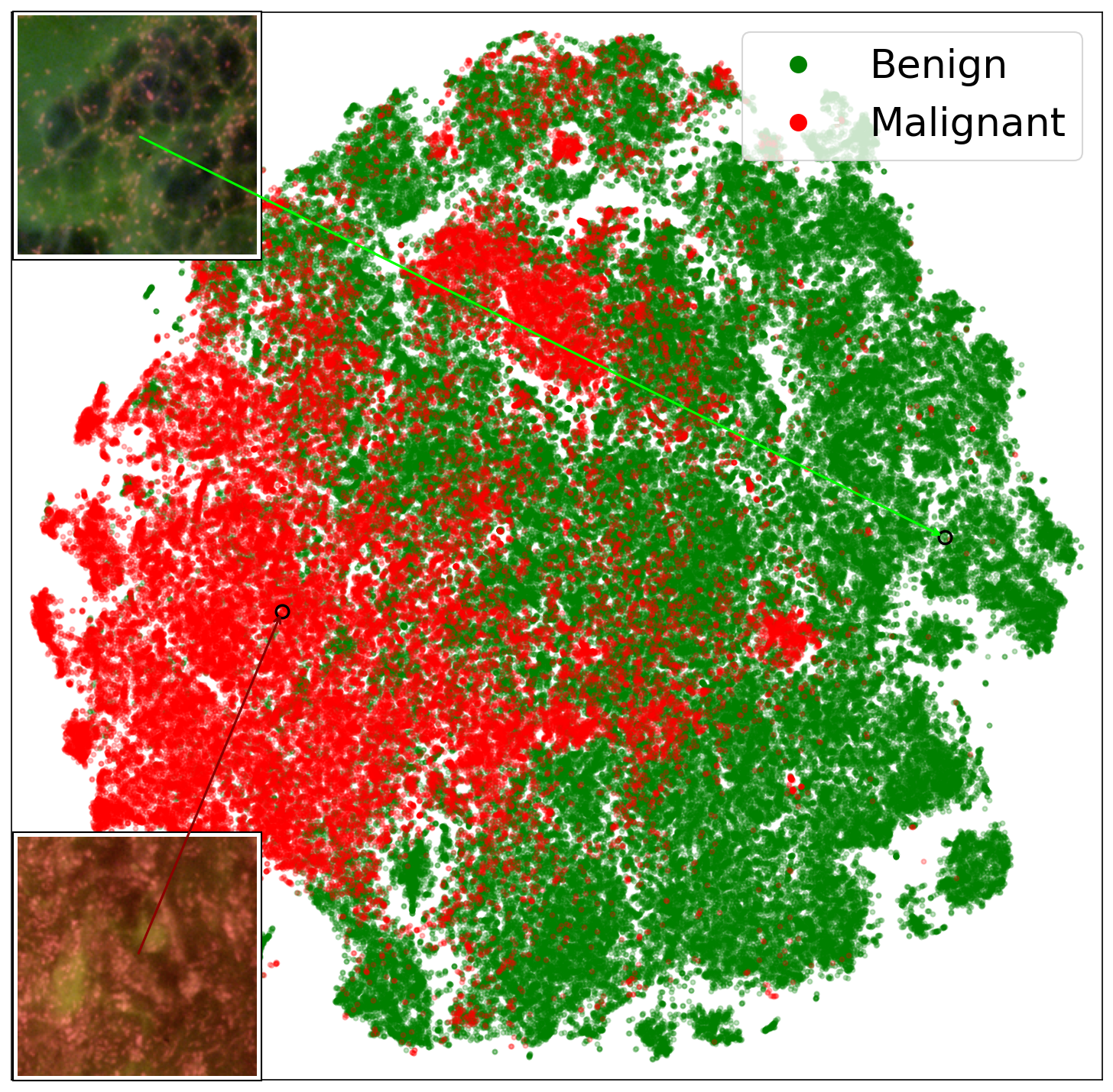}
        \caption{Real Features}
    \end{subfigure}
    \hfill
    \begin{subfigure}[b]{0.33\textwidth}
        \centering
        \includegraphics[width=\textwidth]{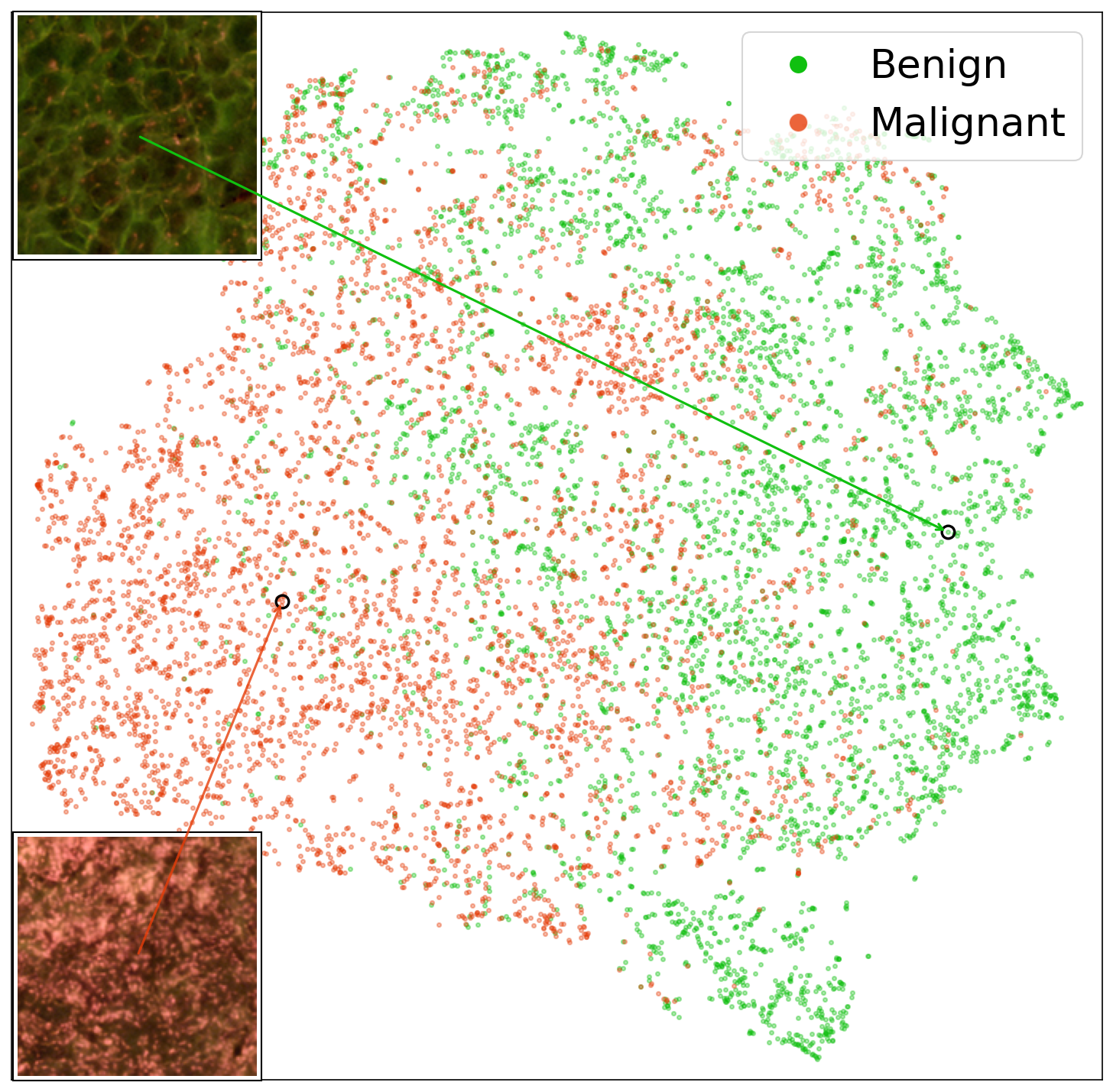}
        \caption{SSL-Guided Synthetic Features}
    \end{subfigure}
    \hfill
    \begin{subfigure}[b]{0.33\textwidth}
        \centering
        \includegraphics[width=\textwidth]{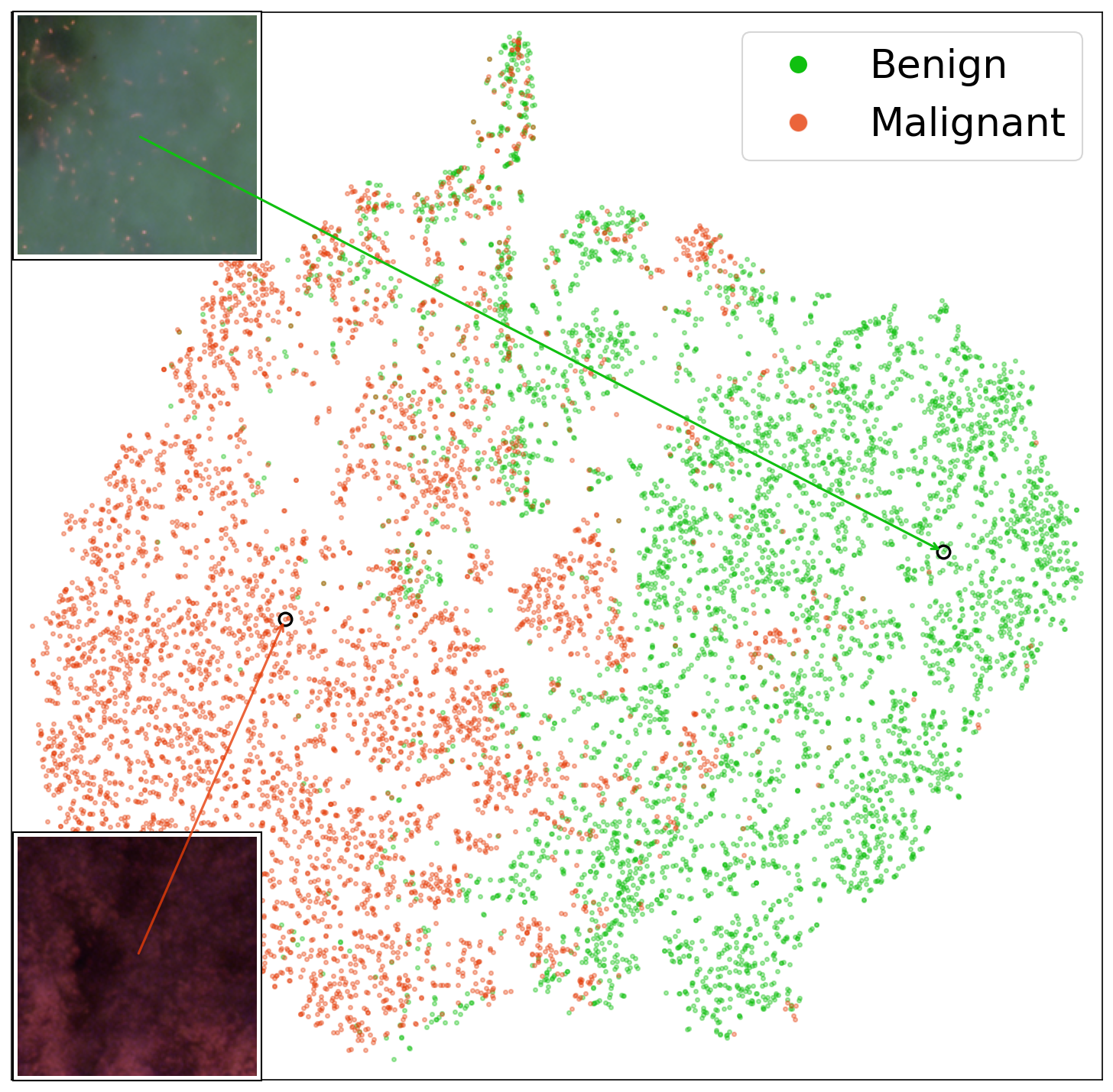}
        \caption{Class Cond. Synthetic Features}
    \end{subfigure}
    \caption{Visualization of feature distributions for real and synthetic DUV WSI patches. SSL-guided synthesis better resembles the true feature distribution, producing sharper, more detailed images, whereas class-conditioned synthesis yields blurrier, less intricate results. Moreover, its feature space resembles the real feature space better than class-guided synthetic features. }
    \label{fig:visualization}
\end{figure*}
\subsection{Dataset and Experimental Setup}
The dataset includes 142 DUV WSIs (58 benign, 84 malignant) from the Medical College of Wisconsin. A total of 172,984 non-overlapping 400×400 patches were extracted (48,619 malignant, 124,365 benign) with labels from pathologist annotations. {DINO was fine-tuned for 30 epochs on the DUV Patch-Level dataset using a ViT-Base backbone with a batch size of 64 per GPU. The student network was trained with AdamW, a learning rate of 5$\times 10^{-4}$, and an initial weight decay of $0.04$. The teacher network used a momentum of 0.996 and a temperature of 0.04.} Both the LDM and its VAE were fine-tuned on the dataset using pretrained checkpoints from \cite{b14}. The U-Net was optimized with AdamW at a learning rate of $2.5 \times 10^{-6}$ for 10 epochs. Sampling employed a classifier-free guidance scale of 2 with 50 DDIM steps. SSL-based patch synthesis followed \cite{b17}, using the fine-tuned VAE suggested in \cite{b16}. DDPM was fine-tuned on the dataset using checkpoints from \cite{b26}.

\subsection{Visual Inspection}
Figure \ref{fig:visualization} shows t-SNE plots \cite{b23} of real patches, SSL-guided synthetic patches, and class-conditioned synthetic patches, along with example images. SSL-guided patches closely match the feature distribution of real patches, revealing clearer cellular and tissue structures in both benign and malignant categories. This means they better match the real data and capture realistic details. Class-conditioned patches, on the other hand, are blurrier, sometimes occupy regions in feature space not covered by real patches, and show only general category information, so they resemble real features less. Overall, SSL embeddings provide more detailed and meaningful features. The Fréchet Inception Distance (FID) \cite{b24} also improves from 54.63 (class) to 45.72 (SSL), showing that SSL-guided synthesis produces more realistic patches.
\begin{table}[t]
\centering
\small 
\setlength{\tabcolsep}{3pt} 
\caption{WSI-level classification performance comparison with PatchViT~\cite{b4} using 5-fold CV.}
\label{table1}
\begin{tabular}{lccc}
\toprule
\textbf{Synthesis} & \textbf{Accuracy} & \textbf{Sensitivity} & \textbf{Specificity} \\
\midrule
N/A & 94.31$\pm$2.94 & 94.04$\pm$0.26 & 94.55$\pm$7.27 \\
DDPM + Class Cond. & 95.05$\pm$2.88 & 95.29$\pm$2.37 & 94.55$\pm$7.27 \\
LDM + Class Cond. & 95.05$\pm$2.88 & 95.29$\pm$2.37 & 94.55$\pm$7.27 \\
LDM + SSL Emb Cond. & \textbf{96.47$\pm$0.13} & \textbf{96.46$\pm$2.90} & \textbf{96.36$\pm$4.45} \\
\bottomrule
\end{tabular}
\vspace{-5pt}
\end{table}
\subsection{Classification Performance}
Table~\ref{table1} shows the WSI-level classification results using 5-fold cross-validation. Adding synthetic patches from the class-conditioned LDM and DDPM slightly improved performance to $95.05\%$ accuracy and $95.29\%$ sensitivity. The best results were achieved with the SSL-conditioned LDM ($96.47\%$ accuracy, $96.46\%$ sensitivity, $96.36\%$ specificity). The improvement is notable, even though the added synthetic patches made up only $5\%$ of the total dataset. {The similar results of the class-conditioned DDPM and LDM suggest that class labels alone do not provide enough information to capture subtle variations within each class or fine diagnostic details. Since both models use the same coarse-class conditioning, their generated patches capture only broad category features, leading to comparable outcomes despite differences in model architecture. In contrast, SSL embeddings carry richer, more detailed information learned directly from real data, which helps the LDM generate more realistic, varied patches that capture complex tissue structures, leading to better patch-level classification and improved WSI-level performance.}
\vspace{-5pt}
\section{Conclusion}
This study presents a method to address the challenges deep learning models face under data scarcity, specifically for breast cancer classification of DUV WSIs. Realistic DUV WSI patches were generated using an LDM guided by Self-Supervised Learning embeddings, which captured more detailed cellular and tissue structures for both benign and malignant categories, as reflected in the FID score of 45.72. The synthetic patches were combined with real ones to improve the performance of DL models and WSI-level classification. The quantitative results showed that WSI-level accuracy and sensitivity improved to $96.47\%$ and $96.46\%$, showing the effectiveness of the method. The study shows that the proposed method can tackle limited data in the medical domain, especially for DUV WSIs, leading to more accurate and reliable diagnostic models that can support careful margin inspection in Breast-Conserving Surgery and improve patient outcomes.

\section{Compliance with Ethical Standards}
\vspace{-8pt}
Approved by the Institutional Review Board of the Medical College of Wisconsin; all DUV WSI data were de-identified per the Declaration of Helsinki.
\vspace{-10pt}

\bibliographystyle{IEEEbib}
\bibliography{strings,refs}

@Article{b1,
  author  = {To, T. and Lu, T. and Jorns, J. M. and Patton, M. and Schmidt, T. G. and Yen, T. and Yu, B. and Ye, D. H.},
  title   = {Deep learning classification of deep ultraviolet fluorescence images toward intra-operative margin assessment in breast cancer},
  journal = {Frontiers in Oncology},
  year    = {2023},
  volume  = {13},
  pages   = {1179025}
}

@Article{b3,
  author  = {Dosovitskiy, A. and Beyer, L. and Kolesnikov, A. and Weissenborn, D. and Zhai, X. and Unterthiner, T. and Dehghani, M. and Minderer, M. and Heigold, G. and Gelly, S. and Uszkoreit, J. and Houlsby, N.},
  title   = {An image is worth 16x16 words: Transformers for image recognition at scale},
  journal = {arXiv preprint arXiv:2010.11929},
  year    = {2020}
}

@InProceedings{b4,
  author    = {Afshin, Pouya and Helminiak, David and Lu, Tongtong and Yen, Tina and Jorns, Julie M. and Patton, Mollie and Yu, Bing and Ye, Dong Hye},
  title     = {Breast Cancer Classification in Deep Ultraviolet Fluorescence Images Using a Patch-Level Vision Transformer Framework},
  booktitle = {Proceedings of the 47th Annual International Conference of the IEEE Engineering in Medicine and Biology Society (EMBC)},
  year      = {2025},
  pages     = {1--6},
  doi       = {10.1109/EMBC58623.2025.11253275}
}

@InProceedings{b5,
  author    = {Ghahfarokhi, S. S. and To, T. and Jorns, J. M. and Yen, T. and Yu, B. and Ye, D. H.},
  title     = {Deep learning for automated detection of breast cancer in deep ultraviolet fluorescence images with diffusion probabilistic model},
  booktitle = {2024 IEEE International Symposium on Biomedical Imaging (ISBI)},
  year      = {2024},
  publisher = {IEEE}
}

@InProceedings{b6,
  author    = {Li, S. and others},
  title     = {Iterative online image synthesis via diffusion model for imbalanced classification},
  booktitle = {International Conference on Medical Image Computing and Computer-Assisted Intervention},
  year      = {2024},
  publisher = {Springer Nature Switzerland, Cham}
}

@Article{b7,
  author    = {Goodfellow, I. and Pouget-Abadie, J. and Mirza, M. and Xu, B. and Warde-Farley, D. and Ozair, S. and Courville, A. and Bengio, Y.},
  title     = {Generative Adversarial Networks},
  journal   = {Communications of the ACM},
  year      = {2014},
  volume    = {63},
  number    = {11},
  pages     = {139--144},
  publisher = {ACM}
}

@InProceedings{b8,
  author    = {Razavi, A. and Van den Oord, A. and Vinyals, O.},
  title     = {Generating Diverse High-Fidelity Images with VQ-VAE-2},
  booktitle = {Advances in Neural Information Processing Systems (NeurIPS)},
  year      = {2019},
  volume    = {32},
  publisher = {Curran Associates, Inc.}
}

@InProceedings{b12,
  author    = {Shin, H.C. and Tenenholtz, N.A. and Rogers, J.K. and Schwarz, C.G. and Senjem, M.L. and Gunter, J.L. and Michalski, M.},
  title     = {Medical Image Synthesis for Data Augmentation and Anonymization Using Generative Adversarial Networks},
  booktitle = {International Workshop on Simulation and Synthesis in Medical Imaging (SASHIMI)},
  year      = {2018},
  pages     = {1--11},
  publisher = {Springer International Publishing},
  address   = {Cham},
  month     = sep
}

@misc{b13,
  author       = {Ho, J. and Salimans, T.},
  title        = {Classifier-Free Diffusion Guidance},
  howpublished = {arXiv preprint arXiv:2207.12598},
  year         = {2022}
}

@misc{b14,
  author       = {Song, J. and Meng, C. and Ermon, S.},
  title        = {Denoising Diffusion Implicit Models},
  howpublished = {arXiv:2010.02502},
  year         = {2020}
}

@InProceedings{b15,
  author    = {Rombach, R. and Blattmann, A. and Lorenz, D. and Esser, P. and Ommer, B.},
  title     = {High-Resolution Image Synthesis with Latent Diffusion Models},
  booktitle = {Proceedings of the IEEE/CVF Conference on Computer Vision and Pattern Recognition (CVPR)},
  year      = {2022}
}

@InProceedings{b16,
  author    = {Yellapragada, Srikar and Graikos, Alexandros and Prasanna, Prateek and Kurc, Tahsin and Saltz, Joel and Samaras, Dimitris},
  title     = {PathLDM: Text Conditioned Latent Diffusion Model for Histopathology},
  booktitle = {Proceedings of the IEEE/CVF Winter Conference on Applications of Computer Vision (WACV)},
  month     = {January},
  year      = {2024},
  pages     = {5182--5191}
}

@InProceedings{b17,
  author    = {Graikos, Alexandros and Yellapragada, Srikar and Le, Minh-Quan and Kapse, Saarthak and Prasanna, Prateek and Saltz, Joel and Samaras, Dimitris},
  title     = {Learned Representation-Guided Diffusion Models for Large-Image Generation},
  booktitle = {Proceedings of the IEEE/CVF Conference on Computer Vision and Pattern Recognition (CVPR)},
  month     = {June},
  year      = {2024},
  pages     = {8532--8542}
}

@article{b20,
  author  = {Zhang, Q. and Xiao, J. and Niu, D. and Zhang, Z. and Ding, S. and Li, Z.},
  title   = {Geometry-complete latent diffusion model for 3D molecule generation},
  journal = {Bioinformatics},
  year    = {2025},
  volume  = {41},
  number  = {8},
  pages   = {btaf426}
}

@InProceedings{b21,
  author    = {Caron, M. and Touvron, H. and Misra, I. and J{\'e}gou, H. and Mairal, J. and Bojanowski, P. and Joulin, A.},
  title     = {Emerging properties in self-supervised vision transformers},
  booktitle = {Proceedings of the IEEE/CVF International Conference on Computer Vision (ICCV)},
  year      = {2021},
  pages     = {9650--9660}
}

@article{b22,
  author    = {Ho, J. and Jain, A. and Abbeel, P.},
  title     = {Denoising Diffusion Probabilistic Models},
  journal   = {Advances in Neural Information Processing Systems},
  volume    = {33},
  pages     = {6840--6851},
  year      = {2020},
  url       = {https://proceedings.neurips.cc/paper/2020/hash/4c5bcfec8584af0d967f1ab10179ca4b-Abstract.html}
}

@article{b23,
  author    = {van der Maaten, L. and Hinton, G.},
  title     = {Visualizing data using t-SNE},
  journal   = {Journal of Machine Learning Research},
  volume    = {9},
  number    = {Nov},
  pages     = {2579--2605},
  year      = {2008},
  url       = {https://www.jmlr.org/papers/volume9/vandermaaten08a/vandermaaten08a.pdf}
}

@article{b24,
  author    = {Heusel, M. and Ramsauer, H. and Unterthiner, T. and Nessler, B. and Hochreiter, S.},
  title     = {GANs trained by a two time-scale update rule converge to a local Nash equilibrium},
  journal   = {Advances in Neural Information Processing Systems},
  volume    = {30},
  year      = {2017}
}

@inproceedings{b26,
  author    = {Nichol, Alexander Quinn and Dhariwal, Prafulla},
  title     = {Improved Denoising Diffusion Probabilistic Models},
  booktitle = {International Conference on Machine Learning},
  series    = {PMLR},
  volume    = {139},
  pages     = {8162--8171},
  year      = {2021},
  url       = {http://proceedings.mlr.press/v139/nichol21a.html}
}

\end{document}